\documentclass[runningheads]{llncs}
\usepackage{graphicx}
\usepackage{amsmath,amssymb} % define this before the line numbering.
\usepackage{color}
\usepackage{multirow}
\usepackage{cite}
\begin{document}
\title{Boosted Attention: Leveraging Human Attention for Image Captioning} 
% Replace with your title

\titlerunning{Boosted Attention Captioning}
% Replace with a meaningful short version of your title
%
\author{Shi Chen\orcidID{0000-0002-3749-4767} \and
Qi Zhao\orcidID{0000-0003-3054-8934}}
%
%Please write out author names in full in the paper, i.e. full given and family names. 
%If any authors have names that can be parsed into FirstName LastName in multiple ways, please include the correct parsing, in a comment to the volume editors:
%\index{Lastnames, Firstnames}
%(Do not uncomment it, because you may introduce extra index items if you do that, we will use scripts for introducing index entries...)
\authorrunning{S. Chen and Q. Zhao}
% Replace with shorter version of the author list. If there are more authors than fits a line, please use A. Author et al.
%

\institute{Department of Computer Science and Engineering,\\
	University of Minnesota\\
	\email{ \{chen4595,qzhao\}@umn.edu}
}
\maketitle              % typeset the header of the contribution
\begin{abstract}
Visual attention has shown usefulness in image captioning, with the goal of enabling a caption model to selectively focus on regions of interest. Existing models typically rely on top-down language information and learn attention implicitly by optimizing the captioning objectives. While somewhat effective, the learned top-down attention can fail to focus on correct regions of interest without direct supervision of attention. Inspired by the human visual system which is driven by not only the task-specific top-down signals but also the visual stimuli, we in this work propose to use both types of attention for image captioning. In particular, we highlight the complementary nature of the two types of attention and develop a model (Boosted Attention) to integrate them for image captioning. We validate the proposed approach with state-of-the-art performance across various evaluation metrics.

\end{abstract}
\section{Introduction}

Image captioning aims at generating fluent language descriptions on a given image. Inspired by the human visual system, in the past few years, visual attention has been incorporated in various image captioning models~\cite{icml2015_xuc15,NIPS2016_6167,Rennie_2017_CVPR,Lu_2017_CVPR}. Attention mechanisms encourage models to selectively focus on specific regions while generating captions instead of scanning through the whole image, avoiding information overflow as well as highlighting visual regions related to the task.

\begin{figure}
\centering
\includegraphics[width=0.48\textwidth]{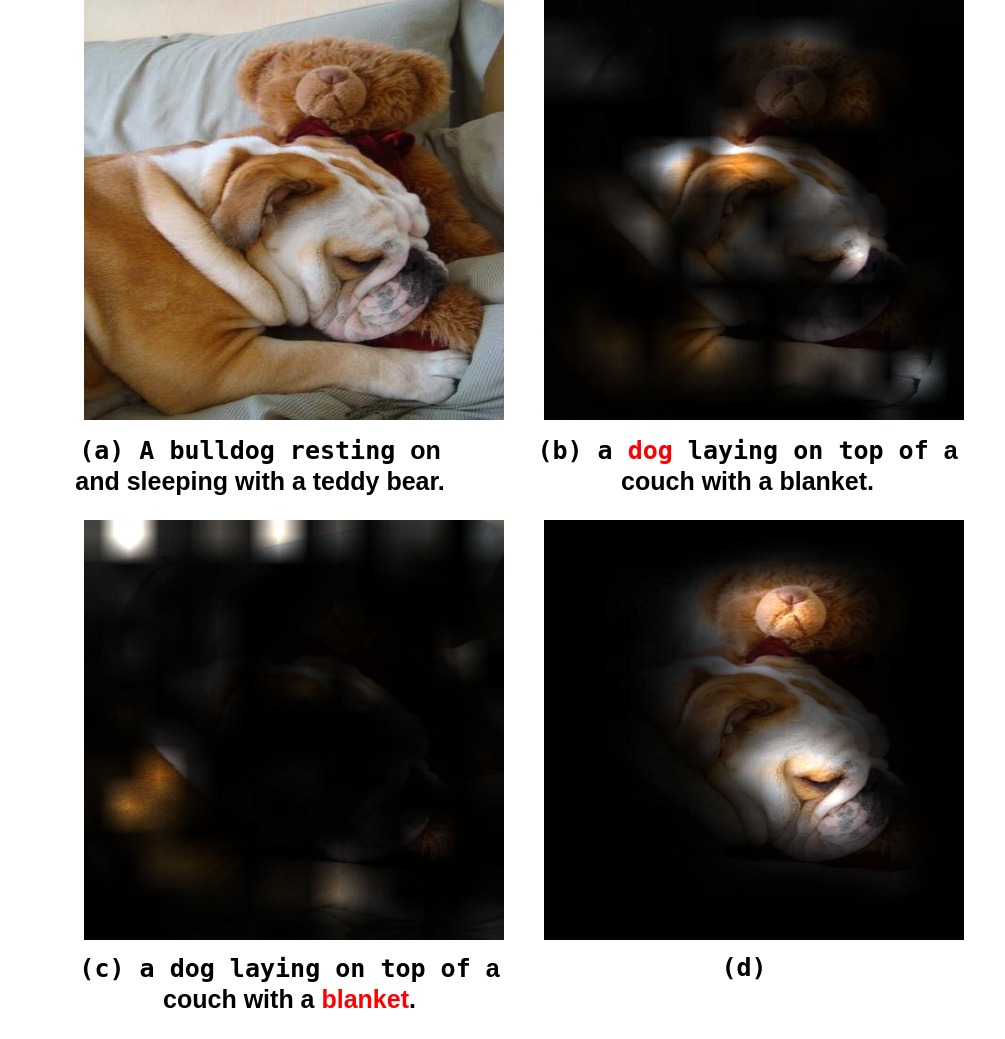}%
\caption{Top-down attention may fail to focus on objects of interest. (a): original image with human-generated caption, (b-c) two top-down attention maps and their corresponding model-generated captions, and (d) stimulus-based attention map for the image. Words related to the top-down attention maps are colored in red.}
\label{intro}
\end{figure}

Following the success made in \cite{icml2015_xuc15}, visual attention in most conventional image captioning models is developed in a top-down fashion on a word basis. That is, visual attention is computed for each generated word based on visual information from the image and the partially generated natural language description. While such mechanism (\textit{i.e.}, top-down attention) aims at connecting natural language and visual content, without prior knowledge on the visual content in terms of salient regions (\textit{i.e.}, stimulus-based attention), the computed visual attention can fail to concentrate on objects of interest and attend to irrelevant regions. As shown in Figure \ref{intro}, a model with only top-down attention focuses on non-salient regions in the background (Figure~\ref{intro}(c)) and does not capture salient objects in the image, \textit{i.e.}, \textit{bulldog} and \textit{teddy bear} according to the human-generated caption.    

Human attention is driven by both task-specific top-down signals and task-independent visual stimuli. For visual tasks such as image captioning, humans would naturally deploy their gaze based on both top-down and stimulus-based information during the exploration. As a result, the objects being mentioned in the same image by different people are largely consistent and correlated with the objects highlighted by the stimulus-based attention \cite{kiwon_cvpr13_gaze}. Therefore, we propose that the visual stimuli can be a reasonable source for locating salient regions in image captioning, which can also complement top-down attention that relates to specific tasks. In Figure \ref{intro}(d), we see that stimulus-based attention successfully attends to regions corresponding to objects of interest as mentioned in the human-generated caption.         

In this work, we conduct qualitative analyses to understand the role of human stimulus-based attention in image captioning. We then present a Boosted Attention method that leverages stimulus-based attention for image captioning. More specifically, we combine the stimulus-based attention with top-down captioning attention to construct a novel attention mechanism that encourages models to attend to visual features based on task-specific top-down signals from natural language while at the same time focusing on salient regions highlighted by task-independent stimulus. Quantitative results on the Microsoft COCO \cite{502} (MSCOCO) and Flickr30K \cite{7410660} datasets show that incorporating stimulus-based attention is able to significantly improve the model performance across various evaluation metrics. We also visualize the results to qualitatively illustrate the complementary role of the two types of attention in image captioning. Our method is general and works with various image captioning models. 

\section{Related Works}
\textbf{Image Captioning.} Generating natural language description based on a still image has gained increasing interest in the recent years. To generate captions, \cite{Farhadi, Kulkarni11babytalk, Li:2011:CSI:2018936.2018962} first extract a set of attributes related to elements within an image and then generate language description based on the detected attributes. Several works \cite{LNCS86920529, Ordonez:2011:im2text, Hodosh:2013:FID:2566972.2566993} view image captioning as a ranking description problem and tackle the problem by conducting a query to retrieve descriptions lies close to an image on embedding space. With the successes of Deep Neural Networks (DNNs), a number of works \cite{DBLP:conf/cvpr/VinyalsTBE15, icml2015_xuc15, 8099611, 7298932, 8100150, 8237362, 8099610} have developed neural network based methods to generate image captions. Typically, these methods use Convolutional Neural Networks (CNNs) as visual encoder to extract visual features and generate captions with Recurrent Neural Networks (RNNs) such as Long Short Term Memory (LSTM) \cite{Hochreiter:1997:LSM:1246443.1246450}. \\

\noindent \textbf{Top-down Attention in Captioning.} Top-down visual attention has been widely used on various image captioning models in order to allow models to selectively concentrate on objects of interest. Xu \textit{et al.} \cite{icml2015_xuc15} combine the memory vector of LSTM with visual features from CNN and feed the fused features to an attention network to compute the weights for features at different spatial locations. Yang \textit{et al.} \cite{NIPS2016_6167} propose a reviewer module that applies the visual attention mechanism for multiple times during generating the next word. In \cite{Lu_2017_CVPR}, an adaptive mechanism is proposed that assigns weights not only to visual features but also to a feature vector obtained based on the memory state of LSTM, since it is unnecessary to attend to the visual features for generating specific words such as `the' and `a'. Besides applying the attention mechanism on the spatial domain, Chen \textit{et al.} \cite{8100150} introduce channel-wise attention which is operated on different filters within a convolutional layer. Most of these models generate visual attention in a top-down fashion using the original visual features and top-down language information from the partially generated caption. Without direct supervision or prior knowledge with stimulus-based attention from the images, however, the computed top-down attention can fail to concentrate on the correct objects of interest and attend to irrelevant background. \\

\noindent \textbf{Stimulus-based Attention in Captioning.} To boost the performance of image captioning models, a few works attempt to use human stimulus-based attention. Sugano \textit{et al.} \cite{DBLP:journals/corr/SuganoB16} utilize ground truth human gaze to split top-down attention for gazed and non-gazed regions. Cornia \textit{et al.} \cite{8026277} integrate human attention in a captioning model similar as \cite{DBLP:journals/corr/SuganoB16} but replace the human gaze with predicted saliency maps. In \cite{8237534}, Tavakoli \textit{et al.} analyze the effects on stimulus-based attention in captioning by substituting the top-down attention with stimulus-based attention. While these models suggest that human attention can have positive effects on image captioning, they either incorporate only stimulus-based attention or use stimulus-based attention to separate the top-down attention at different locations, resulting in relatively marginal improvement over corresponding baselines. 

In this work, we propose a Boosted Attention method that incorporates stimulus-based human attention with existing top-down visual attention. While also using human attention, our method differs from the aforementioned works in the following aspects: 1) Different from \cite{8237534} which solely relies on stimulus-based attention, we emphasize that it is necessary to integrate stimulus-based attention with top-down attention. 2) Unlike \cite{DBLP:journals/corr/SuganoB16, 8026277} which utilize stimulus-based attention to split top-down attention and extract features from regions either attended by both attention (gazed) or not attended by stimulus-based attention (non-gazed), our method extracts features from regions attended by either attention so both contribute directly with an equal role, naturally enabling the two types of attention to complement each other. Experimental results validate the complementary nature of them, which contributes to the significant boost in captioning performance. 3) Instead of using the spatial map for encoding stimulus-based attention like \cite{8237534, DBLP:journals/corr/SuganoB16, 8026277}, we integrate the attention via attentional CNN features. Compared to spatial map, our features encode more abundant information and introduce channel-wise attention in addition to spatial attention.

\section{The role of Stimulus-based Attention in Image Captioning}\label{corr}
Though human-generated captions are relatively free-form, and with considerable inter-subject variance in descriptions, there exists a large degree of agreement in what people describe (\textit{i.e.}, mentioned words in the captions) and what people look (\textit{i.e.}, fixated objects with stimulus-based attention). In this section, we explore the role of stimulus-based attention in image captioning. Specifically, we show the correlations between stimulus-based attention and captioning attention by comparing them on the SALICON \cite{7298710} dataset under different evaluation metrics. Note that to provide insights on how stimulus-based attention could contribute to the captioning task, the captioning attention we use here is derived from ground truth labels from MSCOCO and seen as ground truth attention for generating the captions.

\begin{figure}
\centering
\includegraphics[width=0.55\textwidth]{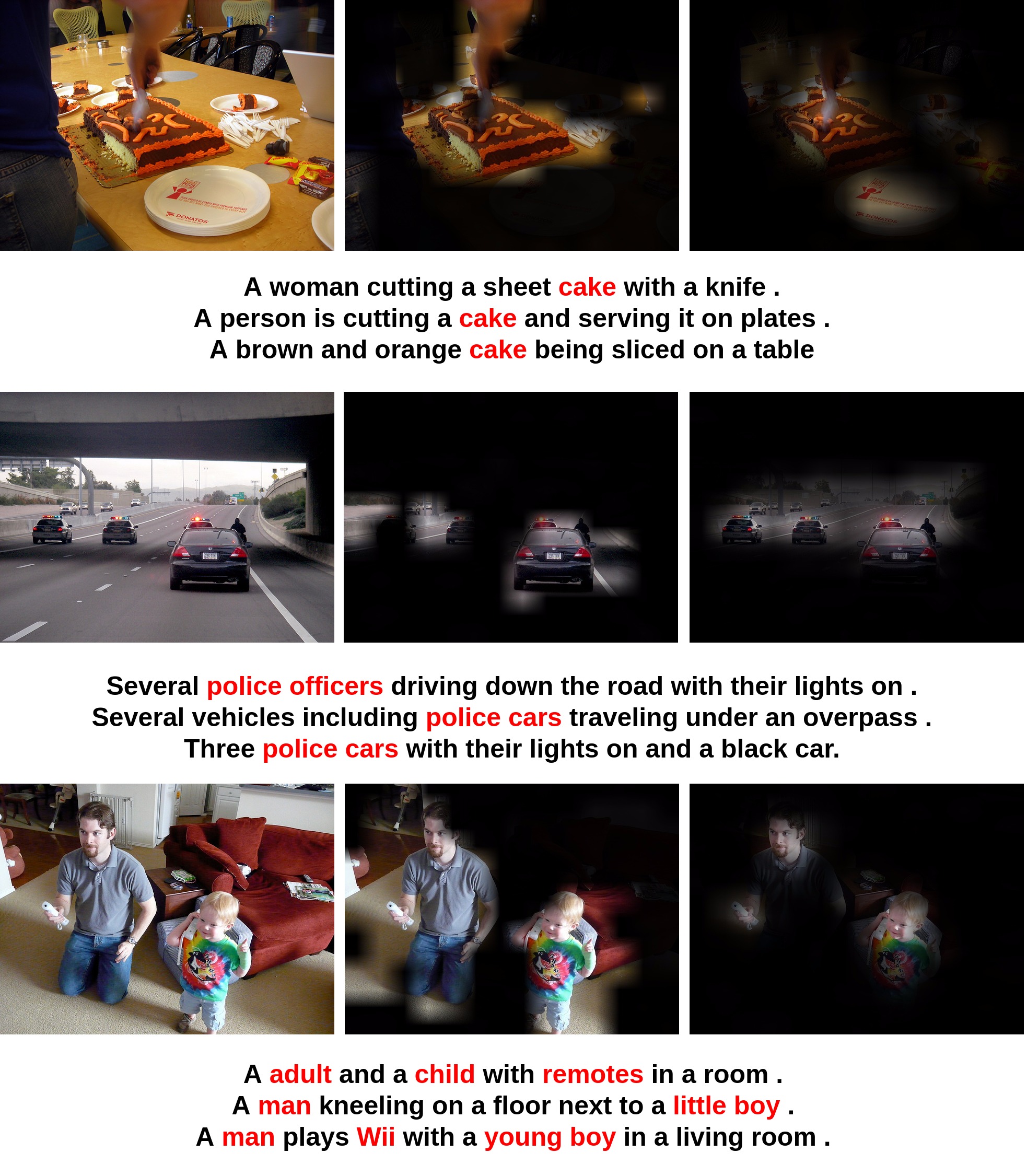}%
\caption{Visualization for image captioning attention and stimulus-based attention. From left to right: original images, ground truth image captioning attention maps, stimulus-based attention maps. Captions are shown at the bottom of the images with objects of interest mentioned in multiple captions highlighted by red color.}
\label{map_compare}
\end{figure}

Similar to \cite{8237534}, we generate captioning attention using visual object category to sentence's noun (VOS) mapping (please refer to the supplementary materials for details). The evaluation metrics used in the comparison include Coefficient Correlation (CC), Spearman's Rank Correlation (Spearman) and Similarity (SIM) \cite{7410387}. Additionally, we also compute the probability of objects being described given that they are fixated by stimulus-based attention, \textit{i.e.}, $P(d|f)$. To compute this probability, we first set up a small threshold (\textit{i.e.}, $0.1$) to filter out the false positive introduced during map re-scaling, then traverse all saliency fixations within the captioning attention map. For each fixation, if the attention value is above the predefined threshold, we consider at that fixation the corresponding object is mentioned in the captions. 

Quantitative evaluations show that the objects described in the captions are likely to be fixated by stimulus-based attention with the probability $P(d|f) = 0.465$. According to \cite{8237534}, the probability of an object being mentioned given that it exists (\textit{i.e.} $P(d|e)$) is around 0.2, thus stimulus-based attention increases the probability of selecting objects of interest by more than 2, providing reasonably good prior knowledge of the objects of interest for image captioning. However, note that since stimulus-based attention commonly attends to only parts of the salient objects instead of covering all or sometimes even majority of the pixels in the objects, the correlations between stimulus-based attention and captioning attention are not high, with $CC = 0.222$, $SIM = 0.353$ and $Spearman = 0.324$. Thus, even though stimulus-based attention is capable of partially capturing objects of interest for image captioning, solely relying on stimulus-based attention may not be sufficient for an image captioning model. Figure \ref{map_compare} shows examples of captioning attention and corresponding stimulus-based attention. We see that stimulus-based attention, while correctly locating objects of interest (\textit{i.e.}, \textit{cake}, \textit{police car}, \textit{man}, \textit{remote} and \textit{boy}), it typically covers part of the salient regions displayed in the captioning attention maps.      

\section{Boosted Attention Method}
As mentioned in section \ref{corr}, on the one hand, objects of interest in stimulus-based attention are reasonably consistent with objects of interest in image captioning, suggesting that stimulus-based attention can be used to provide prior knowledge for image captioning. On the other hand, however, with certain level of discrepancy, in both location and coverage, stimulus-based attention alone could lead to loss of visual information and thus decreasing the quality of generated captions.

We therefore propose a Boosted Attention method for image captioning that incorporates stimulus-based attention into the conventional top-down attention framework of a captioning model. The stimulus-based attention is combined with top-down attention to construct a new attention mechanism called Boosted Attention, which encourages the model to focus on certain visual features based on top-down language signals while at the same time attending to the salient regions highlighted by the stimulus-based attention. In all of our experiments, the stimulus-based attention is obtained from a pre-trained saliency prediction network and details about the network can be found in section \ref{exp}.

\begin{figure*}
\centering
\includegraphics[width=0.7\textwidth]{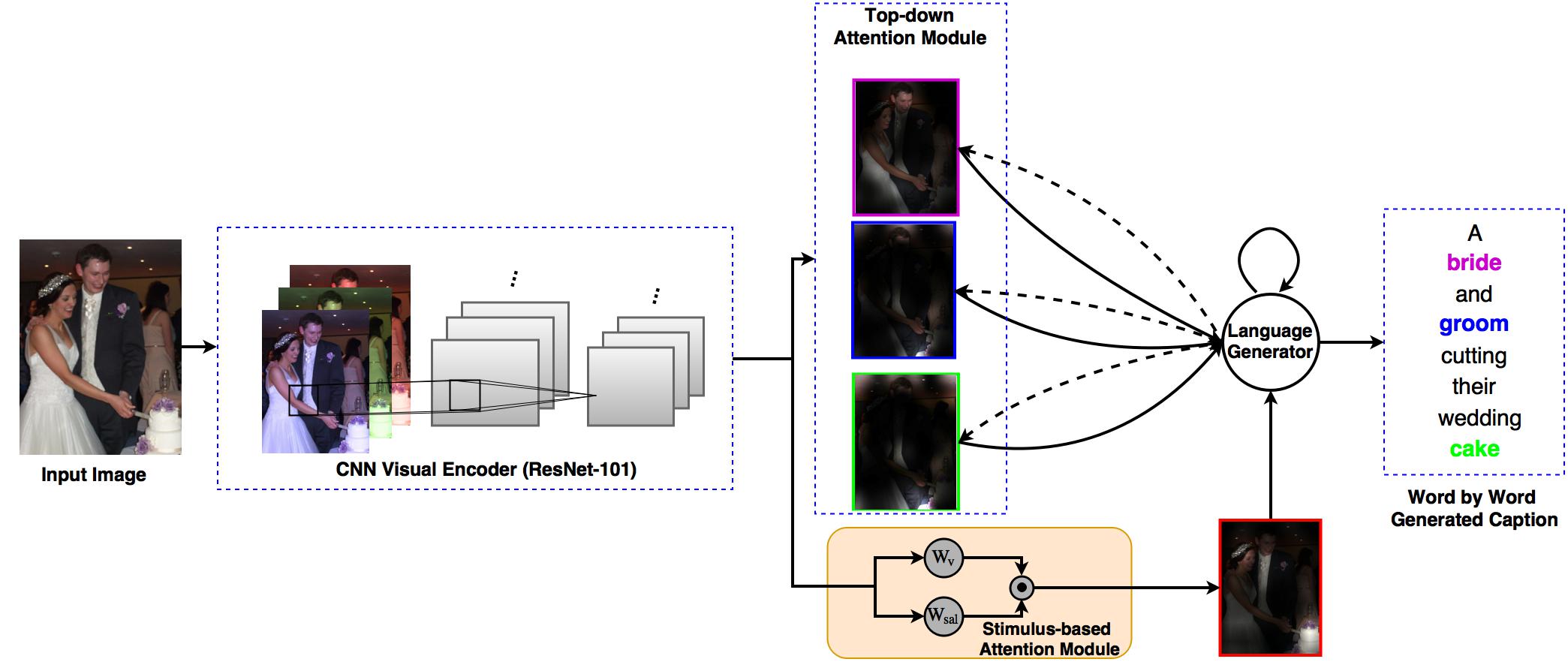}%
\caption{An illustration of architecture design for proposed Boosted Attention method. Top-down attention maps and their corresponding words are highlighted in purple, blue, green color, while stimulus-based attention map is shown in the red frame.}
\label{architecture}
\end{figure*}

Figure \ref{architecture} illustrates the high-level architecture of our method. The model first takes a single raw image as input and encodes it with a CNN Visual Encoder to obtain the visual features. The encoded features are then passed through a Top-down Attention Module and our Stimulus-based Attention Module in parallel, computing the top-down attention and integrating stimulus-based attention. The proposed Stimulus-based Attention Module mainly consists of three parts, a convolutional layer $W_{sal}$ pre-trained on saliency prediction for producing the stimulus-based attention features (attentional CNN features, section \ref{attn_feat}), a convolutional layer $W_{v}$ that further encodes the visual features, and an integration module $\odot$ that combines stimulus-based attention and visual features. After processing with both the Top-down Attention Module and Stimulus-based Attention Module, visual features integrated with two attention are fed into the Language Generator to sequentially produce the caption. 

Note that the proposed method is general and works with different top-down attention and language generation algorithms (\textit{i.e.}, the Top-down Attention Module and the Language Generator in Figure \ref{architecture}). Details about the modules depend on a selected baseline model and the ones used in this work are described in section \ref{exp}.  

\subsection{Attentional CNN Features}\label{attn_feat}
Instead of using the final output of the saliency prediction network (\textit{i.e.}, the saliency map), we propose to make use of features from intermediate layers of the network which could encode richer information about stimulus-based attention. In this section, we formulate and provide intuitions behind using the attentional CNN features to encode stimulus-based attention. 

Considering a fully-convolutional saliency prediction network, we denote it as the equation below (for simplicity we only take the last two layers into considerations):
\begin{equation} \label{feat_map}
 S = softmax(W_{m} \ \delta(W_{sal} I))   
\end{equation}

\noindent where $I$ is the output of previous layers with ReLU activation, $W_{sal}$ and $W_{m}$ represent weight parameters in the layer that is used to produce attentional CNN features and output saliency map respectively, $\delta$ denotes the ReLU activation and $S$ is the saliency map. The kernel size of both convolutional layers is 1, which enables the model to better capture cross-filter correlations as discussed in \cite{DBLP:journals/corr/abs-1709-01507}.  

As shown in Equation \ref{feat_map}, $W_{sal}$ here constructs both channel-wise attention and spatial attention. Specifically, with the use of ReLU activation ensures non-negativity, to highlight salient regions in the saliency map $W_{sal}$ needs to construct the correlations between filters and stimulus-based attention (\textit{i.e.} suppressing filters that have negative correlations and emphasizing those have positive correlations). These correlations (channel-wise attention) are determined by the signs and magnitude of weights in $W_{m}$, \textit{e.g.}, negative weights lead to decrease of activation in S and thus indicate negative correlations, larger weights emphasize more significant contributions. Furthermore, due to the use of spatial softmax activation, $W_{sal}$ also considers the correlations between features and stimulus-based attention on spatial domain, resulting in the spatial attention. 

Therefore, we in this work use $W_{sal}$ to produce attentional CNN features for encoding stimulus-based attention, constructing not only spatial attention widely used in various captioning models but also channel-wise (filter-wise) attention \cite{8100150} that recently found beneficial for image captioning. In Figure \ref{qualitative} we visualize attention maps computed with the CNN features, the results demonstrate that attentional CNN features utilized by our model are capable of highlighting various regions of interest. 

\subsection{Integrating Stimulus-based Attention} \label{integrate_sec}
This section discusses our integration method for introducing stimulus-based attention. We first integrate stimulus-based attention with visual features using an asymmetric function as follows:

\begin{equation} \label{integrate}
 I^{'} = W_{v}I \circ log(W_{sal}I + \epsilon)
\end{equation}

\noindent where I and $I^{'}$ are the visual features before and after integrating stimulus-based attention, $W_{v}$ represents weights in an additional convolutional layer that further encodes the visual features and $W_{sal}$ is the same as in Equation \ref{feat_map}, $\circ$ denotes hadamard product, and $\epsilon$ is a hyper-parameter. Note that $\odot$ in Figure \ref{architecture} denotes the whole integration process of Equation \ref{integrate}.

The intuitions behind this integration method are three-fold: First, $W_{v}$ further encodes visual features, allowing them to adapt to the cross-filter correlations with stimulus-based attention that are stored in $W_{sal}$. Second, by introducing logarithm, we aim at alleviating the effects of co-adaptation between $W_{v}$, $W_{sal}$ and smoothing the contributions of stimulus-based attentional features. Third, with the hyper-parameter $\epsilon$ we form a residual mechanism, preserving the original information in visual features and thus preventing potential information loss caused by applying stimulus-based attention. This mechanism is crucial in the proposed integration method, because stimulus-based attention alone may fail to attend to all regions of interest and it is reasonable to allow the model to extract features attended by either one of the attention (stimulus-based or top-down). In our experiments, we define $\epsilon$ as a mathematical constant $e$ to preserve the identity of the original visual features. Additional discussion on selecting the hyper-parameter is provided in the supplementary materials.

After obtaining the visual features attended by the stimulus-based attention (\textit{i.e.} $I^{'}$), we apply top-down attention on them via hadamard product, enabling two attention to complement to each other. That is, when stimulus-based attention fails to attend to some regions of interest, top-down attention can attend to those regions via assigning larger weights, and vice versa. We further study the corporation between the two types of attention in section \ref{att_corpor}.  

\section{Experiments} \label{exp}
\noindent \textbf{Dataset and Evaluation.} We evaluate our method on two popular datasets: 1) Microsoft COCO \cite{502}, where most images contain multiple objects in complex natural scenes with abundant context information. The dataset includes 82783, 40504, 40775 images for training, validation and online evaluation, each has 5 corresponding captions. We use the publicly available Karapthy's split \cite{7298932} for both training and offline evaluation. 2) Flickr30K \cite{7410660}, where most images depict human performing various activities. It has a total of 31000 images from Flickr, each has 5 corresponding captions. Due to the lack of official split, in order to compare with other works we follow split from \cite{7298932}. Four automatic metrics are used for evaluation, including BLEU \cite{Papineni:2002:BMA:1073083.1073135}, ROUGEL \cite{rouge}, METEOR \cite{Lavie:2007:MAM:1626355.1626389} and CIDEr \cite{DBLP:conf/cvpr/VedantamZP15}. \\

\noindent \textbf{Saliency Prediction Network.} In order to integrate stimulus-based attention, we construct a saliency prediction network with 2 convolutional layers (note that features from the last convolutional layer of a ResNet-101 are viewed as inputs). The first convolutional layer has $2048$ filters while the second layer projects the CNN features to spatial saliency map using a single filter. The kernel size for both layers is set as $1$ and the whole saliency network can be represented as Equation \ref{feat_map}. We optimize the model on SALICON dataset with cross-entropy loss and SGD optimizer using learning rate $2.5 \times 10^{-4}$. Batch size is set to 1. Weights from the first layer of saliency prediction network is utilized to initialize the stimulus-based attention module in the proposed method (\textit{i.e.} $W_{sal}$ in Equation \ref{integrate}).\\

\noindent \textbf{Baseline Model.} To demonstrate the effectiveness of our method and the advantages of integrating stimulus-based attention, we apply the proposed method on our baseline model constructed based on Soft Attention \cite{icml2015_xuc15} and several recent tips \cite{8100150, Rennie_2017_CVPR} to enhance performance: we replace the VGG \cite{DBLP:journals/corr/SimonyanZ14a} based visual encoder with a more powerful ResNet-101 \cite{DBLP:journals/corr/HeZRS15} based one. Instead of fine-tuning the encoder, we directly adopt visual features from the last convolutional layer of the visual encoder as input. When extracting the features, no cropping or re-scaling is applied to the original images, instead, an adaptive spatial average pooling layer is utilized to produce features with a fixed size of $2048 \times 14 \times 14$. Unlike \cite{icml2015_xuc15} which trains the model solely on cross-entropy loss, we use the optimization method proposed in \cite{Rennie_2017_CVPR} which contains both supervised learning and reinforcement learning. The LSTM hidden size, word and attention dimensions are set as $512$ in our baseline. The other settings remain the same as the original Soft Attention model. \\

\noindent \textbf{Training.} We train our models following the same settings from \cite{Rennie_2017_CVPR}: we use ADAM \cite{DBLP:journals/corr/KingmaB14} optimizer for training all of the models and batch size is set as 50. Models are first trained on cross-entropy loss under supervised learning framework, with initial learning learning rate $5 \times 10^{-4}$ and Scheduled Sampling \cite{Bengio:2015:SSS:2969239.2969370} feedback probability being 0. During supervised learning, the learning rate is decayed by a factor of $0.8$ every 3 epochs and feedback probability increased by $0.05$ every 5 epochs. After 25 epochs of supervised learning, we further optimize the models under reinforcement learning framework on the CIDEr metric as \cite{Rennie_2017_CVPR}. The initial learning rate for reinforcement learning is set as $5 \times 10^{-5}$ and also decayed by $0.8$ every 3 epochs. In supervised learning we fix the weights for stimulus-based attention ($W_{sal}$ in Equation \ref{integrate}) to establish correlations between filters within parallel layers ($W_{sal}$ and $W_{v}$ in Equation \ref{integrate}), while later on in reinforcement learning we fine-tune stimulus-based attention since the filter correlations have already been established.

\subsection{Quantitative Results}

In this section, we report quantitative results to demonstrate the effectiveness of the proposed method. We perform inter-model comparisons of the proposed method and 8 state-of-the-art models including Soft Attention \cite{icml2015_xuc15}, ATT \cite{7780872}, SCA-CNN \cite{8100150}, SCN-LSTM \cite{8099610}, RLE \cite{8099611}, AdaATT \cite{Lu_2017_CVPR}, Att2all \cite{Rennie_2017_CVPR} and PG-BCMR \cite{8237362}. We also conduct intra-model comparisons on results with and without the proposed approach (\textit{i.e.,} integrating the stimulus-based attention) and whether using pre-trained stimulus-attention for integration. During evaluation, beam search is utilized for generating the captions and the beam size is set as 3. Table \ref{offline_res} and Table \ref{online_res} show the result comparison on Flickr30K and MSCOCO (Karpathy's test split \cite{7298932} and online testing platform).  

According to the comparative results, the proposed Boosted Attention method leads to significant performance increase across all evaluation metrics compared to the original baselines without stimulus-based attention. On Flickr30K, using our method results in 2.6$\%$, 5.6$\%$, 2.3$\%$ and 12$\%$ of relative improvements on BLEU-4, ROUGE-L, METEOR and CIDEr, while on MSCOCO the improvements are 5.7$\%$, 2.0$\%$, 2.7$\%$ and 5.6$\%$ for corresponding evaluation metrics. Moreover, boosted by the stimulus-based attention, our models are capable of achieving state-of-the-art performance on both datasets.

\begin{table}
    \centering
    \small
    \resizebox{0.85\textwidth}{!}{
    \begin{tabular}{c|c c c c|c c c c}
        \hline
        \multirow{2}{*}{Model} & \multicolumn{4}{|c}{Flickr30K} & \multicolumn{4}{|c}{MSCOCO} \\
        \cline{2-9}
         & B@4 & MT & RG & CD & B@4 & MT & RG & CD \\
        \hline
        Soft Attention \cite{icml2015_xuc15} & 0.191 & 0.185 & - & - & 0.243 & 0.239 & - & - \\
        
        ATT \cite{7780872} & 0.230 & 0.189 & - & - & 0.304 & 0.243 & - & -  \\
        
        SCA-CNN \cite{8100150}  & 0.223 & 0.195 & 0.449 & 0.447 & 0.311 & 0.250 & 0.531 & 0.952 \\
        
        SCN-LSTM \cite{8099610} & 0.265 & 0.218 & - & - & 0.330 & 0.257 & - & 1.012 \\

        RLE \cite{8099611} & - & - & - & - & 0.304 & 0.251 & 0.525 & 0.937 \\ 
    
        AdaATT \cite{Lu_2017_CVPR} & 0.251 & 0.204 & 0.467 & 0.531 & 0.332 & 0.266 & 0.549 & 1.085 \\
        
        Att2all \cite{Rennie_2017_CVPR} & - & - & - & - & 0.342 & 0.267 & 0.557 & 1.140 \\
       \hline
        ours-Baseline  & 0.267 & 0.197 & 0.471 & 0.523 & 0.335 & 0.258 & 0.551 & 1.062 \\
        
        ours-BAM$^{*}$  & 0.270 & 0.204 & 0.477 & 0.571 & 0.350 & 0.262 & 0.559 & 1.111 \\
        
        ours-BAM  & 0.274 & 0.208 & 0.482 & 0.586 & 0.354 & 0.265 & 0.562 & 1.122 \\
        \hline
        Improvement ($\%$) & 2.6$\%$ & 5.6$\%$ & 2.3$\%$ & 12.0$\%$ & 5.7$\%$ & 2.7$\%$ & 2.0$\%$ & 5.6$\%$ \\
        \hline
       
    \end{tabular}
    }
    \caption{Performance comparison with the state-of-the-art on Flickr30K and MSCOCO (test split in \cite{7298932}). Baseline is our augmented baseline model without stimulus-based attention, BAM indicates the proposed Boosted Attention model and BAM$^{*}$ denotes the model without using pre-trained stimulus-based attention but with the same architecture as BAM. Reported scores are BLEU-4 (B@4), METEOR (MT), ROUGE-L (RG) and CIDEr (CD). The relative improvement by using the proposed method over its baseline is shown in percentage.}
    
    \label{offline_res}
\end{table}

\begin{table*}
\centering
\resizebox{0.85\textwidth}{!}{
\begin{tabular}{c c c c c c c c}
\hline
 & BLEU-1 & BLEU-2 & BLEU-3 & BLEU-4 & ROUGEL & METEOR & CIDEr \\
\hline
ATT$^{\dagger}$ \cite{7780872} & 0.731 & 0.565 & 0.424 & 0.316 & 0.535 & 0.250 & 0.953 \\

SCA-CNN \cite{8100150} & 0.712 & 0.542 & 0.404 & 0.302 & 0.524 & 0.244 & 0.912 \\

SCN-LSTM$^{\dagger}$ \cite{8099610} & 0.740 & 0.575 & 0.436 & 0.331 & 0.543 & 0.257 & 1.003 \\

PG-BCMR \cite{8237362} & 0.754 & 0.591 & 0.445 & 0.332 & 0.550 & 0.257 & 1.013	 \\

AdaATT$^{\dagger}$ \cite{Lu_2017_CVPR} & 0.748 & 0.584 & 0.444 & 0.336 & 0.550 & 0.264 & 1.042 \\

Att2all$^{\dagger}$ \cite{Rennie_2017_CVPR} & 0.781 & 0.619 & 0.470 & 0.352 & 0.563 & 0.270 & 1.147  \\

ours-BAM$^{\dagger}$ & 0.794 & 0.622 & 0.470 & 0.349 & 0.560 & 0.264 & 1.083 \\  
\hline
\end{tabular}
}
\caption{Online results (C5) on the MSCOCO evaluation platform, $\dagger$ indicates ensemble of models. Our result is obtained from an ensemble of 4 models trained under different random seeds.}
\label{online_res}
\end{table*}   

To further study the contributions of stimulus-based attention, we conduct experiments using a model with the same architecture as the proposed model but not initialized on pre-trained weights for stimulus-based attention . In this case, the stimulus-based attention $W_{sal}$ is trained end-to-end and not fixed during supervised learning. As shown in Table \ref{offline_res}, models with pre-trained stimulus-based attention (BAM) are able to consistently outperform models without stimulus-based attention (BAM$^{*}$), indicating that stimulus-based attention plays an essential role on boosting the performance and the improvement of our method is not merely due to advantages of modifications on architecture. 

\subsection{Qualitative Results} \label{quali}

\begin{figure*}
\centering
\includegraphics[width=0.8\textwidth]{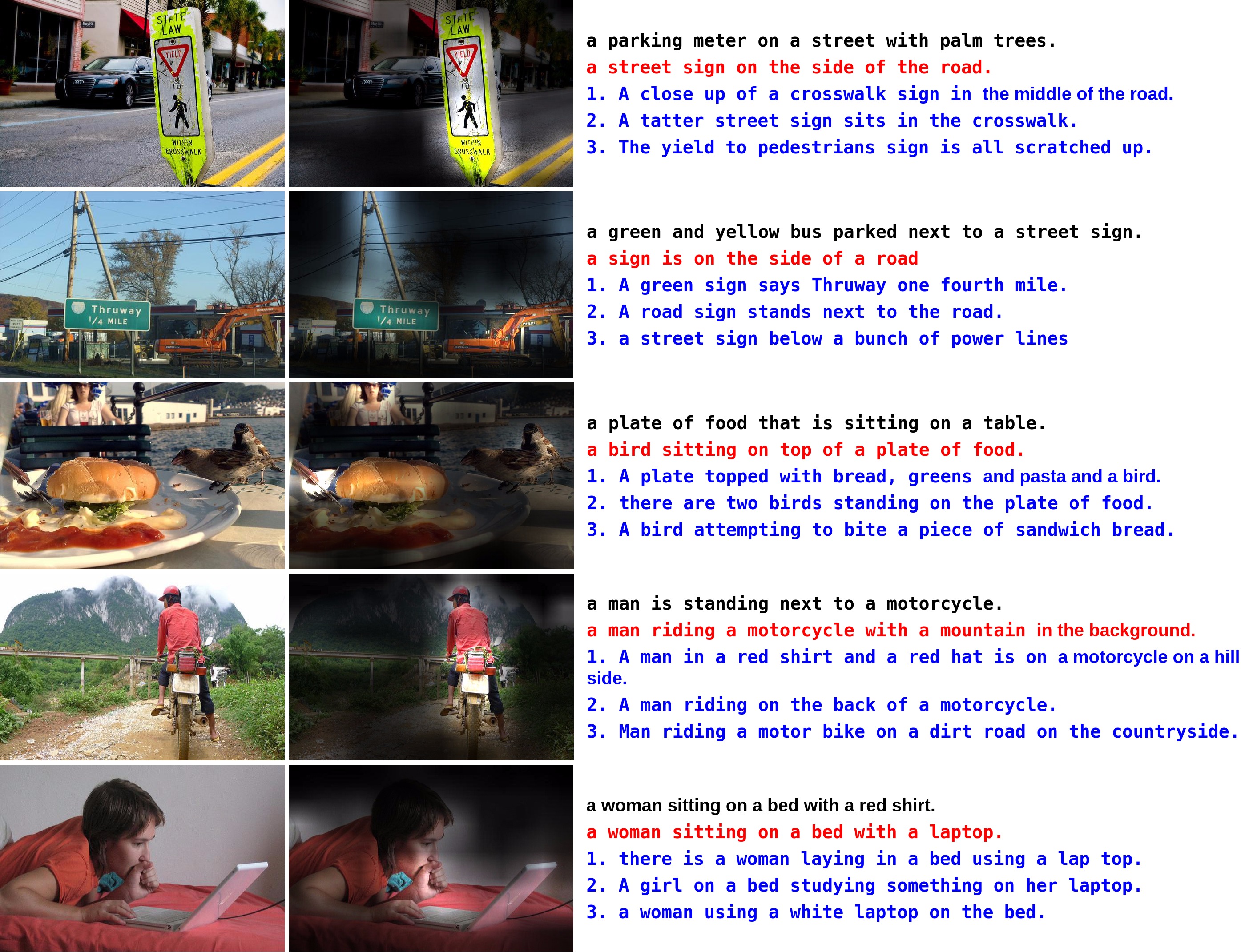}%
\caption{Qualitative results for models with and without using the Boosted Attention method. From left to right: original images, stimulus-based attention map, and captions corresponding to the images. Captions generated by models with and without using Boosted Attention method are colored in red and black respectively, while the ground-truth human generated captions are colored in blue.}
\label{qualitative}
\end{figure*}

In addition to quantitative evaluations, in this section we further demonstrate the effectiveness of proposed method via comparing qualitative results computed by models with and without using our method. Figure \ref{qualitative} shows the captions generated based the two models, together with the corresponding stimulus-based attention maps computed by models using the Boosted Attention method. Stimulus-based attention maps are generated by normalizing the average activation within the CNN features at different spatial locations. 

According to the results, introducing stimulus-based attention helps the model efficiently locate the objects of interest within the visual scenarios and generate better captions. For example, in the top two images, the model using Boosted Attention successfully focuses on the \textit{street signs} similar as humans do (as shown in the attention maps as well as the captions in red) while the model without incorporating stimulus-based attention gets lost in the background objects such as the \textit{palm trees} and \textit{bus} (see the captions in black). Furthermore, the results also indicate that the model with the proposed Boosted Attention method is capable of capturing multiple salient objects within images. For example, for the bottom three images, by incorporating stimulus-based attention, the model is able to concentrate on objects including the \textit{bird}, \textit{mountain} and \textit{laptop} (see the attention maps and captions in red). These objects are missing in the captions generated by the model without using Boosted Attention (captions in black) but mentioned in multiple human generated captions (captions in blue). 

\subsection{Attention Corporation in Image Captioning} \label{att_corpor}

To explore how the two types of attention, \textit{i.e.}, stimulus-based attention and top-down model attention, corporate with each other during the caption generation process, we first evaluate the correlations between the attention maps from the two types of attention. The stimulus-based attention map is extracted using the same method described in section \ref{quali}. Since top-down attention maps are generated for each corresponding word within a caption, we compute the average correlations between stimulus-based attention map and top-down attention maps for different words.

We compute the correlations on the 5000 images from Karpathy's test split \cite{7298932}. Two evaluation metrics commonly used for estimating correlations between spatial maps, \textit{i.e.} Coefficient Correlation (CC) and Spearman's Rank Correlation (Spearman), are utilized for analysis. According to the experimental results, CC and Spearman scores are negative ($CC = -0.256, Spearman = -0.369$), indicating that stimulus-based attention tends to focus on regions different from top-down attention thus the two can potentially complement each other.  

\begin{figure*}
\centering
\includegraphics[width=0.95\textwidth]{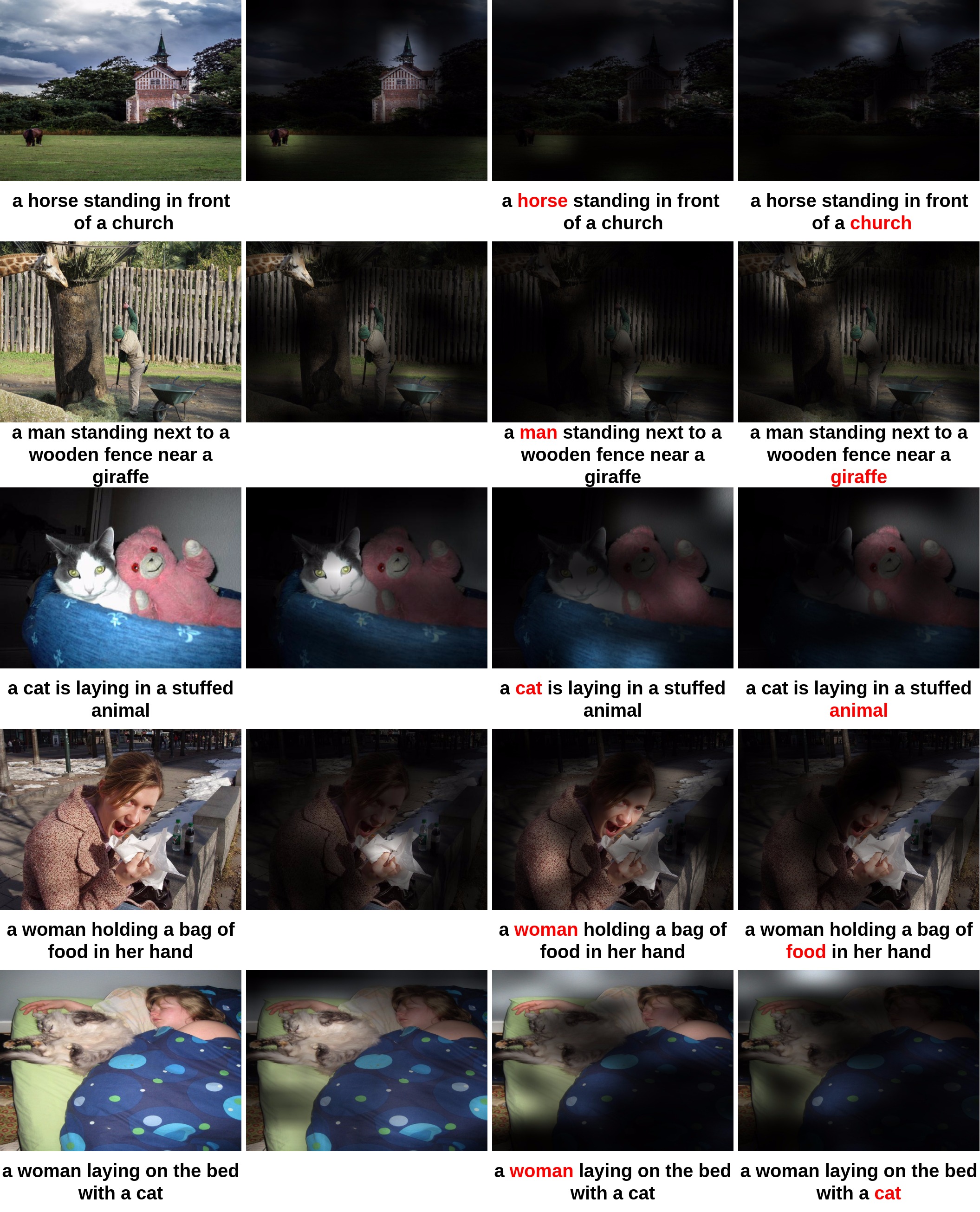}%
\caption{Qualitative results illustrating that the two types of attention complement each other in various situations. From left to right: original images with generated captions, stimulus-based attention maps, top-down model attention maps for different words within the captions. The word associated with a specific top-down attention map is highlighted in red color.}
\label{supp_fig}
\end{figure*}

Next, we show qualitative results to demonstrate that two attention corporate in a complementary manner. Figure \ref{supp_fig} compares top-down attention and its corresponding stimulus-based attention, three typical scenarios for the corporation between attention are summarized as follows: \\

\noindent \textbf{Scenario I}: Stimulus-based attention has successfully captured all of the objects of interest corresponding to generated caption. In this case, top-down attention tends to play a minor role on discriminating the salient regions related to the task. As shown in the first two images, since stimulus-based attention has already concentrated on the objects of interest mentioned in the captions (\textit{i.e.}, \textit{horse} and \textit{church} in the first image, \textit{man} and \textit{giraffe} in the second image), when generating the words corresponding to the objects, top-down attention either does not have a clear focused region (the 1st image) or attends to similar regions as stimulus-based attention (the 2nd image). \\

\noindent \textbf{Scenario II}: Stimulus-based attention concentrates on only part of an object but not covering the entire object (\textit{e.g.}, the 3rd image), or it covers some but not all objects of interest (\textit{e.g.}, the 4th image). Under these situations, top-down attention will focus on the missing regions to enhance the objects of interest and complement stimulus-based attention. In the 3rd image, stimulus-based attention highlights the \textit{cat} but only the bottom part of the \textit{stuffed animal}, therefore in order to collect enough visual information when generating the word \textit{`animal'}, top-down attention is placed on the upper part of the \textit{stuffed animal}. Furthermore, in the 4th image we can see that since stimulus-based attention does not quite focus on the \textit{woman}, during generating the word \textit{`woman'} top-attention significantly emphasizes the face of the \textit{woman} and reveals the lost visual information. \\

\noindent \textbf{Scenario III}: Stimulus-based attention fails to distinguish salient objects with irrelevant background. In this case, top-down attention will play a major role in extracting regions corresponding to the objects of interest. As shown in the 5th image, due to the complexity of the visual scenario, stimulus-based attention confuses the objects of interest (\textit{i.e.} \textit{woman} and \textit{cat} according to the caption) with background objects such as bed and blanket. As a result, the model relies on top-down attention to filter out the irrelevant information and concentrate on regions related to the word being generated.  

\section{Conclusion}
In this work, we propose a Boosted Attention method that leverages human stimulus-based attention to improve the performance of image captioning models. Stimulus-based attention provides prior knowledge on salient regions within the visual scenarios and plays a complementary role to the top-down attention computed by the image captioning models. Experimental results on the MSCOCO and Flickr30K datasets show that the proposed method leads to significant improvements in captioning performance across various evaluation metrics and achieves state-of-the-art results. The proposed method is also general and compatible with various image captioning models using top-down visual attention.

\section*{Acknowledgements}
This work is supported by NSF Grant 1763761 and University of Minnesota Department of Computer Science and Engineering Start-up Fund (QZ).
\clearpage
\bibliographystyle{splncs04}
% \bibliography{egbib}

%
\end{document}